\pdfoutput=1

\documentclass[11pt]{article}

\usepackage[preprint]{acl}

\usepackage{times}
\usepackage{latexsym}

\usepackage[T1]{fontenc}

\usepackage[utf8]{inputenc}

\usepackage{microtype}

\usepackage{inconsolata}

\usepackage{graphicx}
\usepackage{amsmath}
\usepackage{amssymb}
\usepackage{mathtools}
\usepackage{amsthm}
\usepackage{bbm}
\usepackage{algorithm}
\usepackage{algorithmic}
\usepackage{enumitem}
\usepackage{multirow}
\usepackage{booktabs}
\usepackage[most]{tcolorbox}
\usepackage{fontawesome}
\usepackage[normalem]{ulem}
\useunder{\uline}{\ul}{}

\title{AdaptThink: Reasoning Models Can Learn When to Think}

\author{%
  Jiajie Zhang, Nianyi Lin, Lei Hou, Ling Feng, Juanzi Li\\ 
  Tsinghua University
}

\begin{document}

\maketitle

\begin{abstract}
Recently, large reasoning models have achieved impressive performance on various tasks by employing human-like deep thinking. However, the lengthy thinking process substantially increases inference overhead, making efficiency a critical bottleneck. In this work, we first demonstrate that \textit{NoThinking}, which prompts the reasoning model to skip thinking and directly generate the final solution, is a better choice for relatively simple tasks in terms of both performance and efficiency. Motivated by this, we propose \textit{AdaptThink}, a novel RL algorithm to teach reasoning models to choose the optimal thinking mode adaptively based on problem difficulty. Specifically, \textit{AdaptThink} features two core components: (1) a constrained optimization objective that encourages the model to choose \textit{NoThinking} while maintaining the overall performance; (2) an importance sampling strategy that balances \textit{Thinking} and \textit{NoThinking} samples during on-policy training, thereby enabling cold start and allowing the model to explore and exploit both thinking modes throughout the training process. Our experiments indicate that \textit{AdaptThink} significantly reduces the inference costs while further enhancing performance. Notably, on three math datasets, \textit{AdaptThink} reduces the average response length of DeepSeek-R1-Distill-Qwen-1.5B by 53\% and improves its accuracy by 2.4\%, highlighting the promise of adaptive thinking-mode selection for optimizing the balance between reasoning quality and efficiency. Our codes and models are available at \url{https://github.com/THU-KEG/AdaptThink}.
\end{abstract}

\section{Introduction}

\begin{figure}[htbp]
    \centering
    \includegraphics[width=\linewidth]{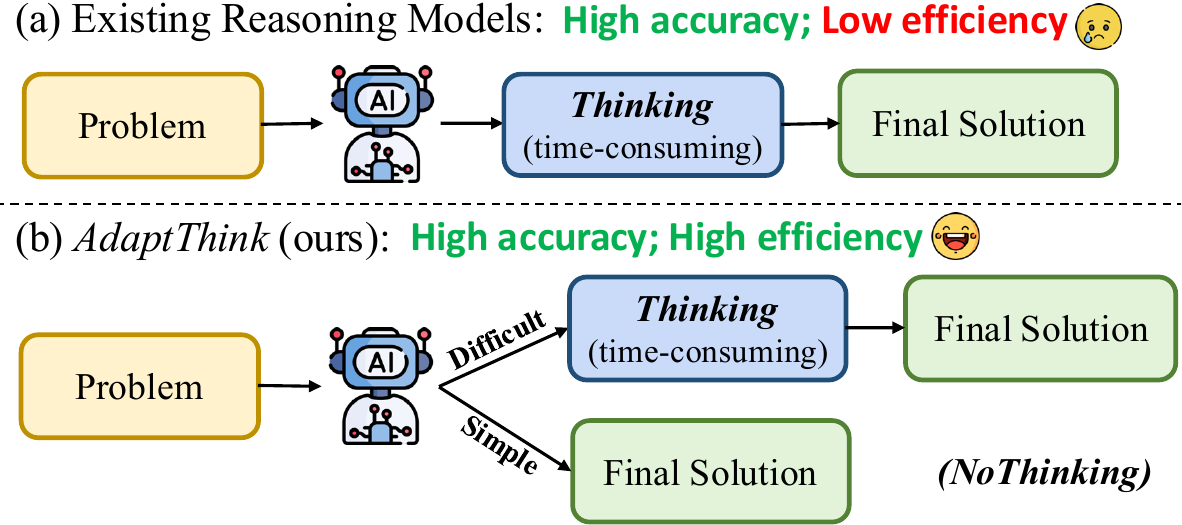}
    \caption{\textit{AdaptThink} enables models to adaptively select between \textit{Thinking} or \textit{NoThinking} mode based on problem difficulty, thereby improving reasoning efficiency while further improving overall performance.}
    \label{fig:intro}
\end{figure}

Recent advancements in large reasoning models, such as OpenAI o1\cite{openai2024reasoning} and DeepSeek-R1 \cite{deepseekai2025deepseekr1}, have demonstrated remarkable capabilities in tackling complex tasks. Given a problem, these models first engage in a long chain of thought—also referred to as \textit{Thinking}—where they iteratively explore different approaches, accompanied by reflection, backtracking, and self-verification. Subsequently, they produce a final solution that contains only the correct steps and the answer to present to the user. While the long-thinking process markedly enhances the model’s reasoning capacities, it also substantially increases inference overhead and latency~\cite{Qu2025survey, Sui2025}. In particular, for some simple queries where users expect fast, near-instant responses, these models often generate excessive thinking with unnecessarily detailed steps or redundant attempts, resulting in a suboptimal user experience~\cite{chen2025overthink, shen2025dast}.

Existing efforts to improve reasoning efficiency primarily focus on reducing the length of model responses, either through incorporating length-based rewards in reinforcement learning (RL)~\cite{arora2025training, kimiteam2025kimik15}, finetuning with preference pairs that penalizes longer responses~\cite{chen2025overthink, shen2025dast, luo2025o1pruner}, or by merging reasoning and non-reasoning models~\cite{wu2025unlocking}. Nevertheless, these methods still apply thinking to all instances, regardless of whether thinking itself is necessary for every problem. In this work, we draw inspiration from the recently introduced \textit{NoThinking} approach~\cite{ma2025reasoning}, which allows reasoning models to skip the thinking process and directly generate the final solution by prompting with a pseudo-thinking process.
Specifically, we further simplify the approach by prompting the model with an empty thinking segment (i.e., ``\texttt{<think>}\texttt{</think>}''). Our pilot study in Section~\ref{sec:motivation} indicates that \textit{NoThinking} achieves comparable or even better performance than \textit{Thinking} on relatively simple problems (up to high-school competition level), while significantly reducing token usage; the benefits of \textit{Thinking} only become pronounced when the problem is difficult enough. 

In light of this observation, we are curious: \textit{Can the reasoning model learn to select \textit{Thinking} or \textit{NoThinking} mode adaptively based on the difficulty of the input problem, thereby achieving more efficient reasoning without sacrificing or even improving performance?} To this end, we propose \textit{AdaptThink}, a novel RL algorithm to teach reasoning models when to think. Specifically, \textit{AdaptThink} features two core components: (1) a constrained optimization objective that encourages the model to choose \textit{NoThinking} while ensuring overall performance does not degrade; (2) an importance sampling strategy that balances \textit{Thinking} and \textit{NoThinking} samples during on-policy training, thereby overcoming the challenge of cold start and allowing the model to explore and exploit both thinking modes throughout the whole training process.

Our experiments demonstrate that \textit{AdaptThink} effectively enables reasoning models to adaptively select the optimal thinking mode based on problem difficulty, leading to substantial reductions in inference cost compared to prior approaches, while consistently enhancing model accuracy.
For instance, on GSM8K, MATH500, and AIME2024, \textit{AdaptThink} reduces the average response length of DeepSeek-R1-Distill-Qwen-1.5B by 50.9\%, 63.5\%, and 44.7\%, and improving its accuracy by 4.1\%, 1.4\%, and 1.6\%, respectively. The remarkable results substantiate the potential of difficulty-adaptive thinking-mode selection as a promising paradigm for advancing the trade-off between reasoning performance and efficiency.

In summary, our key contributions are as follows: (1) We simplify the \textit{NoThinking} approach and demonstrate its advantages over \textit{Thinking} for simpler tasks in terms of both performance and efficiency; (2) We propose \textit{AdaptThink}, a novel RL algorithm that empowers reasoning models to adaptively select the optimal thinking mode adaptively based on problem difficulty, thereby substantially reducing inference costs and further improving performance; (3) We conduct extensive experiments to validate the efficacy of \textit{AdaptThink}.

\begin{figure*}[!t]
    \centering
    \includegraphics[width=\linewidth]{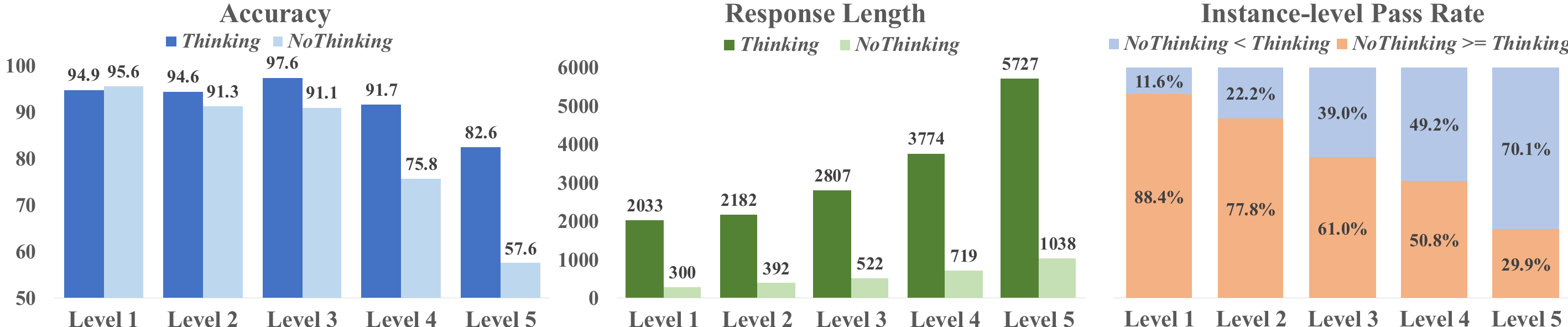}
    \caption{Comparison of DeepSeek-R1-Distill-Qwen-7B using \textit{Thinking} and \textit{NoThinking} mode across different difficulty levels of MATH500 dataset.}
    \label{fig:pilot_sutdy}
\end{figure*}

\section{Related Work}
\paragraph{Large Reasoning Models.} 
Recent frontier large reasoning models (LRMs), such as OpenAI o1~\cite{openai2024reasoning}, DeepSeek-R1~\cite{deepseekai2025deepseekr1}, and QwQ~\cite{qwen2025qwq}, have developed the ability to employ human-like deep thinking in problem solving by generating a long chain of thought before arriving at a final solution. Such advanced ability is typically acquired through large-scale RL with verified rewards or fine-tuning on distilled reasoning traces. Despite promising performance, the lengthy thinking process introduces substantial inference costs and latency. 
Consequently, a variety of approaches have been proposed for more efficient reasoning.

\paragraph{Efficient Reasoning for LRMs.} 
Most existing methods to improve the efficiency of LRMs focus on reducing the token usage in model responses. Some methods incorporate length-based rewards into RL to incentivize more concise responses~\cite{arora2025training, kimiteam2025kimik15} or enable precise control over response length~\cite{aggarwal2025l1}. Other approaches finetune models with length-related preference pairs, which are obtained from best-of-N sampling~\cite{luo2025o1pruner, shen2025dast} or through postprocessing~\cite{chen2025overthink}. Additionally, several works pursue training-free methods to decrease response length, employing techniques such as model merging~\cite{kimiteam2025kimik15, wu2025unlocking} or prompting~\cite{han2024token, muennighoff2025s1, fu2025reasoning, xu2025chainofdraft}. Nevertheless, these methods still utilize long thinking for all problems, while the recent \textit{NoThinking} approach \cite{ma2025reasoning} allows reasoning models to bypass long thinking and directly output the final solution via prompting, achieving performance comparable to \textit{Thinking} in low-token-budget settings. In this work, we further demonstrate that even with a sufficient token budget, \textit{NoThinking} can outperform \textit{Thinking} on simple problems while using significantly fewer tokens. This observation motivates us to propose \textit{AdaptThink} to teach reasoning models to adaptively select the optimal thinking mode based on problem difficulty, which is a new direction for efficient reasoning.

\section{Motivation}

\label{sec:motivation}

\subsection{Preliminary}

Consider a reasoning model parameterized by $\theta$ and denoted by $\pi_\theta$. Given a prompt $x\!=\![x_1, \dots, x_n, \texttt{<think>}]$, where $[x_1, \dots, x_n]$ represents the problem and $\texttt{<think>}$ is the special token to start the thinking process, the model generates a response $y\!=\![y_1, \dots, y_l, \texttt{</think>}, y_{l+2}, \dots, y_m]$. Here, $[y_1, \dots, y_l]$ corresponds to the thinking, which is a long chain of thought consisting of constant exploration, reflection, and self-verification. The token $\texttt{</think>}$ marks the end of thinking. The remaining sequence, $[y_{l+2}, \dots, y_m]$, denotes the final solution, which only includes the correct steps to solve the problem and the final answer. From the perspective of probability theory, the response $y$ is a sample drawn from the conditional probability distribution $\pi_\theta(\cdot|x)$. Since $y$ is generated in an auto-regressive way, the conditional probability $\pi_\theta(y|x)$ can be decomposed as:

\begin{equation}
    \pi_\theta(y|x) = \prod_{t=1}^m\pi_\theta(y_t|x, y_{<t})
\end{equation}

\subsection{\textit{NoThinking} is Better for Simple Problems}
\label{sec:pilot_study}
Current reasoning models, such as OpenAI o1 and DeepSeek-R1, apply long thinking across all problems (denoted as \textit{Thinking} mode). Though enhancing models' reasoning capabilities, the lengthy thinking process often leads to unnecessary computation overhead, especially for some simple problems that can also be solved by non-reasoning models (e.g., GPT-4o and Qwen-2.5-Instruct) without thinking. 
Recently, \citet{ma2025reasoning} proposed \textit{NoThinking} method, which enables reasoning models to bypass long thinking and directly generate the final solution by prompting with a fake thinking process \textit{``Okay,  I think I have finished thinking.\texttt{</think>}''}, and found it is still effective in low-token-budget settings. 
In this work, we further simplify \textit{NoThinking} by providing the models with an empty thinking (i.e., enforcing the first generated token $y_1=\texttt{</think>}$).
Then, we conduct a pilot study to compare \textit{Thinking} and \textit{NoThinking} from the perspective of problem difficulty, with a sufficient token budget (16K).

Specifically, we utilize MATH500~\cite{math500} dataset for the pilot study since its have categorized problems into five difficulty levels. For each problem, we employ DeepSeek-R1-Distill-Qwen-7B to generate 16 responses using \textit{Thinking} and \textit{NoThinking}, respectively. Then we analyze the accuracy, response length, and instance-level pass rate across the five difficulty levels. As illustrated in Figure~\ref{fig:pilot_sutdy}, although the model is trained using long-thinking data, $\textit{NoThinking}$ still achieves accuracy comparable to \textit{Thinking} on relatively simple problems (Level 1 to 3), and even slightly outperforms \textit{Thinking} on the easiest Level-1 problems. Meanwhile, the average length of \textit{NoThinking} responses is significantly shorter than \textit{Thinking} ones. Additionally, compared to \textit{Nothinking}, \textit{Thinking} only improves the instance-level pass rate for less than half of the problems from Level 1 to 4. Overall, these findings indicates that \textit{Thinking} only brings notable benefits for challenging problems, whereas \textit{NoThinking} can be a better choice for simpler questions in terms of both accuracy and efficiency. This motivates us to explore efficient reasoning from a new perspective: teaching the reasoning model to adaptively select \textit{Thinking} or \textit{NoThinking} mode based on problem difficulty, thereby reducing inference costs while maintaining or even improving the overall performance. To this end, we propose \textit{AdaptThink}, a novel RL algorithm that teaches reasoning models when to think.

\section{AdaptThink}

Our \textit{AdaptThink} algorithm consists of two important components: (1) a constrained optimization objective that incentivizes the model to select \textit{NoThinking} mode, while ensuring the overall performance does not decrease; (2) an importance sampling strategy that balances \textit{Thinking} and \textit{NoThinking} samples during on-policy training, thereby enabling cold start and also allowing the model to explore and exploit both thinking modes throughout the entire training process. We will introduce these two components in detail as follows.

\subsection{Constrained Optimization Objective}
Considering that \textit{NoThinking} mode offers a significant advantage over \textit{Thinking} in reasoning efficiency, an ideal selection policy should prefer to choose \textit{NoThinking} as long as the overall performance is not diminished. In other words, we should maximize the probability of generating \textit{NoThinking} responses while ensuring the model's accuracy does not decline. 

Formally, consider a reasoning model $\pi_\theta$ and a dataset $\mathcal{D}$. Let $\pi_{\theta_\text{ref}}$ denote the reference model, which is the initial $\pi_\theta$ and remains unchanged during training. Let $R(x, y, y^*)$ be the reward function (i.e., accuracy in math solving), where $x$, $y$, and $\hat{y}$ denote the prompt, model response, and golden answer, respectively. It returns 0/1 if $y$ is incorrect/correct. For simplicity, we omit $\hat{y}$ and denote the function as $R(x, y)$. Let $\mathbbm{1}(y_1\!=\!\texttt{</think>})$ be the indicator function, which returns 1 if the first token of $y$ is \texttt{</think>} (i.e., $y$ is a \textit{NoThinking} response), otherwise returns 0. Then our optimization objective can be formulated as:
\begin{align}
    \max\ &\mathbb{E}_{x\sim \mathcal{D}, y\sim \pi_\theta(\cdot|x)}\mathbbm{1}(y_1\!=\!\texttt{</think>}) \notag\\
    s.t.\ &\mathbb{E}_{x\sim \mathcal{D}, y\sim \pi_\theta(\cdot|x)}R(x, y) \ge \notag \\
    &\mathbb{E}_{x\sim \mathcal{D}, y'\sim \pi_{\theta_\text{ref}}(\cdot|x)}R(x, y').  
\end{align}

To solve this constrained optimization problem, we incorporate the constraint into the objective as a penalty term, with a penalty weight $\lambda \ge 0$:
\begin{align}
    \max\ &\mathbb{E}_{x\sim \mathcal{D}, y\sim \pi_\theta(\cdot|x), y'\sim \pi_{\theta_\text{ref}}(\cdot|x)}\mathbbm{1}(y_1\!=\!\texttt{</think>}) \notag \\
    &+ \lambda \big(R(x, y) - R(x, y')\big).
\end{align}
By dividing the both side by $\lambda$, letting $\delta=\frac{1}{\lambda}$, and reorganizing the terms about $\pi_{\theta_\text{ref}}$, we have:
\begin{align}
    \max\ &\mathbb{E}_{x\sim \mathcal{D}, y\sim \pi_\theta(\cdot|x)} \mathbbm{1}(y_1\!=\!\texttt{</think>})\cdot\delta \notag \\
        &+ R(x, y) - \mathbb{E}_{y'\sim \pi_{\theta_\text{ref}}(\cdot|x)}R(x, y').
\end{align}
In practice, $\mathbb{E}_{y'\sim \pi_{\theta_\text{ref}}(\cdot|x)}R(x, y')$ can be approximated by pre-sampling before training. Specifically, we sample $K$ responses from $\pi_{\theta_\text{ref}}(\cdot|x)$ for each $x$, and calculate their mean reward:
\begin{equation}
    \bar{R}_\text{ref}(x) = \frac{1}{K}\sum\limits_{i=1}^K R(x, y'^i), \ y'^i\sim \pi_{\theta_\text{ref}}(\cdot|x).
\label{eq:mean_reference_reward}
\end{equation}
Then the optimization objective becomes:
\begin{align}
    \max\ &\mathbb{E}_{x\sim \mathcal{D}, y\sim \pi_\theta(\cdot|x)} \mathbbm{1}(y_1\!=\!\texttt{</think>})\cdot\delta \notag \\
    &+ R(x, y) - \bar{R}_\text{ref}(x).
\end{align}
Since $\mathbbm{1}(y_1\!=\!\texttt{</think>})$ and $R(x, y)$ are not differentiable, we employ policy gradient method to solve this optimization. Specifically, let $\pi_{\theta_\text{old}}$ be a distribution equal to $\pi_\theta$ without gradient update, and define the advantage function: 
$A(x, y) = \mathbbm{1}(y_1\!=\!{\texttt{</think>}})\cdot\delta + R(x, y) - \bar{R}_\text{ref}(x)$. 
Then the objective can be converted into a PPO-style~\cite{ppo} loss without KL penalty:
\begin{align}
    \mathcal{L}(\theta) = &\!-\!\mathbb{E}_{x\sim \mathcal{D}, y\sim \pi_{\theta_\text{old}}(\cdot|x)}\big[\min\big({\textstyle\frac{\pi_\theta(y|x)}{\pi_{\theta_\text{old}}(y|x)}} A(x, y), \notag \\
    &\operatorname{clip}({\textstyle\frac{\pi_\theta(y|x)}{\pi_{\theta_\text{old}}(y|x)}}, 1\!-\!\epsilon, 1\!+\!\epsilon)A(x, y)\big)\big].
\end{align}
Here, $\operatorname{clip}(\cdot)$ denotes the clipping function, which improves the stability of training.

\begin{algorithm*}[!t]
\caption{\textit{AdaptThink}}
\small
\label{alg:AdaptThink}
\textbf{Input:} policy model $\pi_\theta$; dataset $\mathcal{D}$; hyperparameters $K, \delta, \epsilon$ \\
\textbf{Initialize:} reference model $\pi_{\theta_\text{ref}} \gets \pi_\theta$
\begin{algorithmic}[1]
    \STATE Sample $K$ responses $\{y'^i\}_{i=1}^K\sim\pi_{\theta_\text{ref}}(\cdot|x)$ and calculate $\bar{R}_\text{ref}(x)$ for each $x\in \mathcal{D}$ (Equation ~\ref{eq:mean_reference_reward})
    \FOR{$\text{step}=1, \dots, M$} 
        \STATE Update the old policy model $\pi_{\theta_\text{old}}\gets \pi_\theta$ and importance sampling distribution $\pi_\text{IS}$ (Equation~\ref{eq:importance_sampling_distribution})
        \STATE Sample a batch $\mathcal{D}_b$ from $\mathcal{D}$
        \STATE Sample $K$ responses $\{y^i\}_{i=1}^K\sim\pi_\text{IS}(\cdot|x)$ for each $x\in \mathcal{D}_b$ and estimate $\mathcal{L}_\text{AT}(\theta)$ (Equation ~\ref{eq:adaptive_loss}. Half of $y^i$ are \textit{Thinking} responses and the other half are \textit{NoThinking} responses.)
        \STATE Update the policy model $\pi_\theta$ by minimizing $\mathcal{L}_\text{AT}(\theta)$
    \ENDFOR
\end{algorithmic}
\textbf{Output:} $\pi_\theta$ 
\end{algorithm*}

\subsection{Importance Sampling}
At each step of optimizing $\mathcal{L}(\theta)$ using on-policy training, we sample a batch $\mathcal{D}_b$ from the dataset $\mathcal{D}$, and then sample $K$ responses $\{y^i\}_{i=1}^K$ from $\pi_{\theta_\text{old}}(\cdot|x)$ for each $x\in \mathcal{D}_b$ to estimate $\mathcal{L}(\theta)$. However, since the initial $\pi_\theta$ naturally apply \textit{Thinking} across all problems, it is impossible to get \textit{Nothinking} samples from $\pi_{\theta_\text{old}}$ from the training starts (i.e., $\pi_{\theta_\text{old}}(y_1\!=\!\texttt{</think>}|x) \!\approx\! 0$). As a result, the model can only learn from \textit{Thinking} samples and will never generate \textit{NoThinking} responses. 

To solve this cold-start challenge, we employ the technique of importance sampling. Specifically, we define a new distribution $\pi_\text{IS}(\cdot|x)$:
\begin{equation}
    \pi_\text{IS}(y_t\!=\!a|x, y_{<t}) \!=\! \left\{
    \begin{aligned}
    &0.5, \  \text{if}\ t\!=\!1, a\!=\!\texttt{</think>}; \\
    &0.5, \  \text{if}\ t\!=\!1, a\!=\!w_\text{start}; \\
    &\pi_{\theta_\text{old}}(y_t\!=\!a|x, y_{<t}),\  \text{if}\ t\!>\!1.
    \end{aligned}
    \right.
\label{eq:importance_sampling_distribution}
\end{equation}
Here, $w_\text{start}$ is a common word to start long thinking, such as \textit{``Alright''}. During training, we sample responses $\{y^i\}_{i=1}^K$ from $\pi_\text{IS}(\cdot|x)$ instead of $\pi_{\theta_\text{old}}(\cdot|x)$, so that half of the samples in a batch are in \textit{Thinking} mode and the other half are \textit{NoThinking}. This allows the model to learn from both modes from the beginning of training, and finally adaptively be able to select the appropriate mode. Accordingly, our final loss function of \textit{AdaptThink} becomes:
\begin{align}
    \mathcal{L}_\text{AT}(&\theta) = -\mathbb{E}_{x\sim \mathcal{D}, y\sim \pi_\text{IS}(\cdot|x)}\big[\min\big({\textstyle\frac{\pi_\theta(y|x)}{\pi_\text{IS}(y|x)}} A(x, y), \notag \\
    &\operatorname{clip}({\textstyle\frac{\pi_\theta(y|x)}{\pi_\text{IS}(y|x)}}, 1\!-\!\epsilon, 1\!+\!\epsilon)A(x, y)\big)\big].
\label{eq:adaptive_loss}
\end{align}
In addition to enabling cold start, importance sampling also preserves the opportunities for exploration and exploitation across both \textit{Thinking} and \textit{NoThinking} modes during the entire training process. This prevents $\pi_\theta$ from collapsing into one thinking mode forever and completely ignoring the other, even if the latter mode may demonstrate a greater advantage in the future. Finally, we summarize our \textit{AdaptThink} algorithm in Algorithm~\ref{alg:AdaptThink}.

\subsection{A New Perspective to Understand the Loss}
In this subsection, we provide another perspective to understand our loss function $\mathcal{L}_\text{AT}(\theta)$ by comparing the advantage $A(x, y)$ of \textit{Thinking} and \textit{NoThinking} samples from $\pi_\text{IS}(\cdot|x)$. Given a prompt $x$, we denote the average pass rate of \textit{Thinking} and \textit{NoThinking} samples as $\bar{R}_\text{think}(x)$ and $\bar{R}_\text{nothink}(x)$, respectively. Then their average advantages are:
\begin{align}
    &\bar{A}_\text{think}(x)=\bar{R}_\text{think}(x) - \bar{R}_\text{ref}(x), \notag \\
    &\bar{A}_\text{nothink}(x)=\delta + \bar{R}_\text{nothink}(x) - \bar{R}_\text{ref}(x).
\end{align}
Note that the probability of choosing \textit{NoThinking} (i.e., $\pi_\theta(y_1\!=\!\texttt{</think>}|x)$) and \textit{Thinking} (i.e., $\pi_\theta(y_1\!=\!w^*|x)$) are competitive. Therefore, when optimizing $\mathcal{L}_\text{AT}(\theta)$, $\pi_\theta(y_1\!=\!\texttt{</think>}|x)$ will improve only if $\bar{A}_\text{nothink}(x) \!>\! 0$ and $\bar{A}_\text{nothink}(x) \!>\! \bar{A}_\text{think}(x)$, which give us: 
\begin{align}
    &\bar{R}_\text{nothink}(x) + \delta > \bar{R}_\text{ref}(x), \notag \\
    &\bar{R}_\text{nothink}(x) + \delta > \bar{R}_\text{think}(x).
\end{align}
In other words, only when the problem is simple enough such that the accuracy gap between \textit{NoThinking} and \textit{Thinking}, as well as the reference model, is smaller than $\delta$, $\mathcal{L}_\text{AT}(\theta)$ will favor \textit{NoThinking} and encourage $\pi_\theta$ to directly generate the final solution. For more challenging problems where \textit{NoThinking} lags far behind the other two, $\mathcal{L}_\text{AT}(\theta)$ will prioritize performance and guide $\pi_\theta$ to engage in \textit{Thinking} more frequently. Therefore, $\mathcal{L}_\text{AT}(\theta)$ aligns well with our expectation for difficulty-adaptive thinking in Section~\ref{sec:pilot_study}.

\begin{table*}[!t]
\centering
\resizebox{\linewidth}{!}{
    \setlength{\tabcolsep}{8pt}
\begin{tabular}{l|ccc|ccc|ccc|cc}
\toprule
                                                             & \multicolumn{3}{c|}{GSM8K}                                                                 & \multicolumn{3}{c|}{MATH 500}                                                               & \multicolumn{3}{c|}{AIME 2024}                                                               & \multicolumn{2}{c}{Average}                                   \\
\multirow{-2}{*}{Method}                                     & Acc                         & Length                     & $\text{Ratio}_\textit{NT}$     & Acc                         & Length                      & $\text{Ratio}_\textit{NT}$     & Acc                         & Length                       & $\text{Ratio}_\textit{NT}$     & $\Delta$Acc                  & $\Delta$Length                 \\ \midrule
\multicolumn{12}{l}{\textit{DeepSeek-R1-Distill-Qwen-1.5B}}                                                                                                                                                                                                                                                                                                                                                         \\ \midrule
{\color[HTML]{7F7F7F} $\text{Original}_\textit{Thinking}$}   & {\color[HTML]{7F7F7F} 79.0} & {\color[HTML]{7F7F7F} 978} & {\color[HTML]{7F7F7F} 0.0\%}   & {\color[HTML]{7F7F7F} 80.6} & {\color[HTML]{7F7F7F} 4887} & {\color[HTML]{7F7F7F} 0.0\%}   & {\color[HTML]{7F7F7F} 29.4} & {\color[HTML]{7F7F7F} 12073} & {\color[HTML]{7F7F7F} 0.0\%}   & {\color[HTML]{7F7F7F} -}     & {\color[HTML]{7F7F7F} -}       \\
{\color[HTML]{7F7F7F} $\text{Original}_\textit{NoThinking}$} & {\color[HTML]{7F7F7F} 69.8} & {\color[HTML]{7F7F7F} 280} & {\color[HTML]{7F7F7F} 100.0\%} & {\color[HTML]{7F7F7F} 67.2} & {\color[HTML]{7F7F7F} 658}  & {\color[HTML]{7F7F7F} 100.0\%} & {\color[HTML]{7F7F7F} 14.0} & {\color[HTML]{7F7F7F} 2190}  & {\color[HTML]{7F7F7F} 100.0\%} & {\color[HTML]{7F7F7F} -12.7} & {\color[HTML]{7F7F7F} -79.9\%} \\
$\text{DPO}_\textit{Shortest}$                               & 78.3                        & 804                        & 0.0\%                          & 82.4                        & 3708                        & 0.0\%                          & {\ul 30.7}                  & 10794                        & 0.0\%                          & +0.8                         & -17.5\%                        \\
OverThink                                                    & 77.2                        & 709                        & 0.0\%                          & 81.2                        & 4131                        & 0.0\%                          & 28.3                        & 11269                        & 0.0\%                          & -0.8                         & -16.5\%                        \\
DAST                                                         & 77.2                        & 586                        & 0.0\%                          & {\ul 83.0}                  & {\ul 2428}                  & 0.0\%                          & 26.9                        & {\ul 7745}                   & 0.0\%                          & -0.6                         & -42.1\%                        \\
O1-Pruner                                                    & 74.8                        & \textbf{458}               & 0.0\%                          & 82.2                        & 3212                        & 0.0\%                          & 28.9                        & 10361                        & 0.0\%                          & -1.0                         & -33.9\%                        \\
TLMRE                                                        & {\ul 80.7}                  & 863                        & 0.0\%                          & \textbf{85.0}               & 3007                        & 0.0\%                          & 29.2                        & 8982                         & 0.0\%                          & +2.0                         & -25.3\%                        \\
ModelMerging                                                 & 79.7                        & 603                        & 0.0\%                          & 63                          & 2723                        & 0.0\%                          & 18.1                        & 10337                        & 0.0\%                          & -9.4                         & -32.3\%                        \\
$\text{RFT}_\textit{MixThinking}$                            & 76                          & 1077                       & 8.8\%                          & 72.4                        & 4341                        & 33.4\%                         & 25.2                        & 11157                        & 21.0\%                         & -5.1                         & -2.9\%                         \\
\textit{AdaptThink}                                          & \textbf{83.1}               & {\ul 480}                  & 86.9\%                         & 82.0                        & \textbf{1782}               & 76.8\%                         & \textbf{31.0}               & \textbf{6679}                & 40.4\%                         & \textbf{+2.4}                & \textbf{-53.0\%}               \\ \midrule
\multicolumn{12}{l}{\textit{DeepSeek-R1-Distill-Qwen-7B}}                                                                                                                                                                                                                                                                                                                                                           \\ \midrule
{\color[HTML]{7F7F7F} $\text{Original}_\textit{Thinking}$}   & {\color[HTML]{7F7F7F} 87.9} & {\color[HTML]{7F7F7F} 682} & {\color[HTML]{7F7F7F} 0.0\%}   & {\color[HTML]{7F7F7F} 90.2} & {\color[HTML]{7F7F7F} 3674} & {\color[HTML]{7F7F7F} 0.0\%}   & {\color[HTML]{7F7F7F} 53.5} & {\color[HTML]{7F7F7F} 10306} & {\color[HTML]{7F7F7F} 0.0\%}   & {\color[HTML]{7F7F7F} -}     & {\color[HTML]{7F7F7F} -}       \\
{\color[HTML]{7F7F7F} $\text{Original}_\textit{NoThinking}$} & {\color[HTML]{7F7F7F} 85.1} & {\color[HTML]{7F7F7F} 283} & {\color[HTML]{7F7F7F} 100.0\%} & {\color[HTML]{7F7F7F} 80.6} & {\color[HTML]{7F7F7F} 697}  & {\color[HTML]{7F7F7F} 100.0\%} & {\color[HTML]{7F7F7F} 24.2} & {\color[HTML]{7F7F7F} 1929}  & {\color[HTML]{7F7F7F} 100.0\%} & {\color[HTML]{7F7F7F} -13.9} & {\color[HTML]{7F7F7F} -73.6\%} \\
$\text{DPO}_\textit{Shortest}$                               & 85.7                        & 402                        & 0.0\%                          & 91.6                        & 2499                        & 0.0\%                          & 52.5                        & 8699                         & 0.0\%                          & -0.6                         & -29.5\%                        \\
OverThink                                                    & 86.3                        & 426                        & 0.0\%                          & 89.4                        & 2435                        & 0.0\%                          & 53.1                        & 8744                         & 0.0\%                          & -0.9                         & -28.8\%                        \\
DAST                                                         & 86.7                        & 459                        & 0.0\%                          & 89.6                        & {\ul 2162}                  & 0.0\%                          & 45.6                        & \textbf{7578}                & 0.0\%                          & -3.2                         & -33.4\%                        \\
O1-Pruner                                                    & 87.6                        & 428                        & 0.0\%                          & 86.6                        & 2534                        & 0.0\%                          & 49.2                        & 9719                         & 0.0\%                          & -2.7                         & -24.7\%                        \\
TLMRE                                                        & {\ul 88.9}                  & 756                        & 0.0\%                          & {\ul 91.8}                  & 2899                        & 0.0\%                          & {\ul 54.0}                  & 8633                         & 0.0\%                          & +1.0                         & -8.8\%                         \\
ModelMerging                                                 & 88.4                        & 531                        & 0.0\%                          & 72.6                        & 2280                        & 0.0\%                          & 36.9                        & 8624                         & 0.0\%                          & -11.2                        & -25.5\%                        \\
$\text{RFT}_\textit{MixThinking}$                            & 86.2                        & {\ul 365}                  & 66.5\%                         & 84.8                        & 2411                        & 64.8\%                         & 49.4                        & 9969                         & 10.0\%                         & -3.7                         & -28.0\%                        \\
\textit{AdaptThink}                                          & \textbf{91.0}               & \textbf{309}               & 99.6\%                         & \textbf{92.0}               & \textbf{1875}               & 76.6\%                         & \textbf{55.6}               & {\ul 8599}                   & 6.3\%                          & \textbf{+2.3}                & \textbf{-40.1\%}              \\ \bottomrule
\end{tabular} 
}
\caption{Accuracy (Acc), response length (Length), and the ratio of \textit{NoThinking} responses ($\text{Ratio}_\textit{NT}$) of different methods on three math benchmarks. The best and second results are bolded and underlined, respectively. 
}
\label{tab:main}
\end{table*}
\begin{figure*}[!t]
    \centering
    \includegraphics[width=\linewidth]{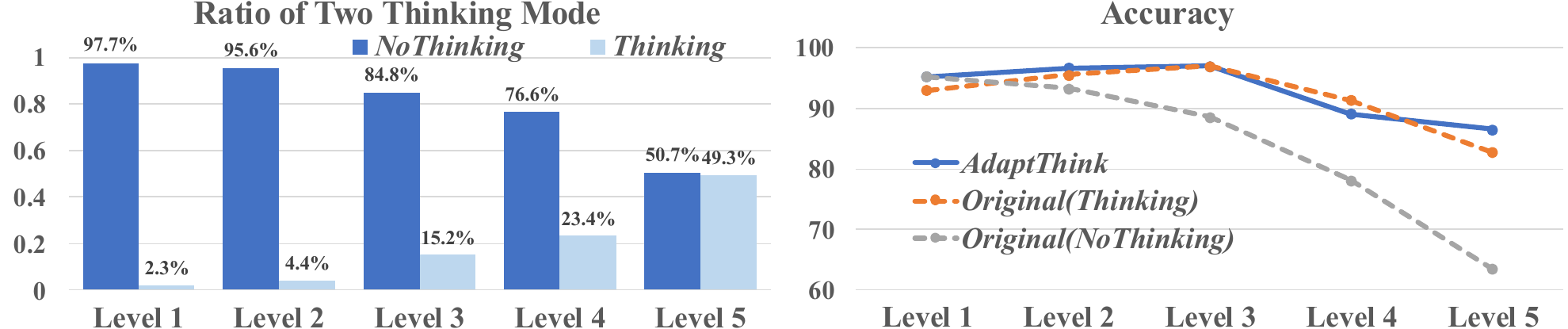}
    \caption{Left: The ratio that \textit{AdaptThink}-7B choose \textit{Thinking} or \textit{NoThinking} across different MATH levels. Right: Comparison of accuracy between \textit{AdaptThink}-7B and DeepSeek-R1-Distill-Qwen-7B using \textit{Thinking} and \textit{NoThinking} across different MATH levels.}
    \label{fig:nothinking_ratio_math500_7B}
\end{figure*}

\section{Experiments}
\subsection{Setup}

\paragraph{Models.} We select DeepSeek-R1-Distill-Qwen-1.5B and DeepSeek-R1-Distill-Qwen-7B, two popular reasoning models that demonstrate impressive performance on math problem solving,  as the initial policy models. 

\paragraph{Dataset and Metrics.} The training dataset we use is DeepScaleR~\cite{deepscaler2025} dataset, which consists of 40K math problems drawn from AIME 1983-2023, AMC, Omni-Math~\cite{omnimath}, and STILL~\cite{STILL}. For evaluation, we use three math datasets with increasing difficulty: GSM8K~\cite{gsm8k} test set (1319 grade school math problems), MATH500~\cite{math500} (500 high-school competition math problems), and AIME 2024 (30 Olympiad-level math problems). For evaluation metrics, we consider both accuracy and response length. 
We also report the average accuracy variation and the average length reduction rate across all the test datasets. 
Considering the limited size of AIME 2024, we repeatedly sample 16 responses for each case and report the average results.  For all models, we set the evaluation context size to 16K, and set the temperature to 0.6 as suggested in DeepSeek's model cards.

\paragraph{Implementation Details.} We build our code based on VeRL~\cite{verl} framework. The training context size, batch size, and the learning rate are set to 16K, 128, and 2e-6, respectively. The hyperparameters $K$, $\delta$, and $\epsilon$ in $\textit{AdaptThink}$ are set to 16, 0.05, and 0.2, respectively. The comparison of using different $\delta$ is shown in Section~\ref{sec:analyses}. We train the models for 1 epoch, which is 314 steps in total. For the 1.5B model, we use one 8$\times$H800 node and cost about 32 hours. For the 7B model, we use four 8$\times$H800 nodes and cost about 28 hours. Finally, we select the checkpoints on 300 and 150 steps for the 1.5B and 7B models, respectively, where the models' accuracy and response lengths achieve a good balance.

\subsection{Baselines}
We compare \textit{AdaptThink} with the following representative methods for efficient reasoning: 

\begin{table*}[!t]
\centering
\resizebox{\linewidth}{!}{
    \setlength{\tabcolsep}{7pt}
\begin{tabular}{l|ccc|ccc|ccc|cc}
\toprule
\multirow{2}{*}{Method} & \multicolumn{3}{c|}{GSM8K}                  & \multicolumn{3}{c|}{MATH500}                & \multicolumn{3}{c|}{AIME 2024}              & \multicolumn{2}{c}{Average}  \\
                        & Acc  & Length & $\text{Ratio}_\textit{NT}$ & Acc  & Length & $\text{Ratio}_\textit{NT}$ & Acc  & Length & $\text{Ratio}_\textit{NT}$ & $\Delta$Acc & $\Delta$Length \\ \midrule
\multicolumn{12}{l}{\textit{DeepSeek-R1-Distill-Qwen-1.5B}}                                                                                                                                                        \\ \midrule
$\text{Original}_\textit{Thinking}$                & 79.0 & 978    & 0.0\%                      & 80.6 & 4887   & 0.0\%                      & 29.4 & 12073  & 0.0\%                      & -           & -              \\
$\text{Original}_\textit{NoThinking}$              & 69.8 & 280    & 100.0\%                    & 67.2 & 658    & 100.0\%                    & 14.0 & 2190   & 100.0\%                    & -12.7       & -79.9\%        \\ 
\midrule
\multicolumn{12}{l}{\textit{AdaptThink-1.5B} }                                                                                                                                                 \\ \midrule
$\delta\!=\!0$          & 84.6 & 718    & 70.4\%                     & 86   & 2511   & 50.0\%                     & 34.8 & 9279   & 0.2\%                      & +5.5        & -32.8\%        \\ 
$\delta\!=\!0.01$       & 82.4 & 638    & 55.6\%                     & 82.4 & 2473   & 67.0\%                     & 32.5 & 9165   & 19.8\%                     & +2.8        & -36.1\%        \\
$\delta\!=\!0.02$       & 83.1 & 628    & 75.7\%                     & 83.4 & 2337   & 62.6\%                     & 31.3 & 8696   & 21.3\%                     & +2.9        & -38.6\%        \\
$\delta\!=\!0.05$       & 83.1 & 480    & 86.9\%                     & 82   & 1782   & 76.8\%                     & 31   & 6679   & 40.4\%                     & +2.4        & -53.0\%        \\
$\delta\!=\!0.075$      & 83.2 & 580    & 71.8\%                     & 80.2 & 1621   & 84.2\%                     & 29.2 & 6087   & 64.2\%                     & +1.2        & -52.4\%        \\
$\delta\!=\!0.1$        & 82.5 & 358    & 91.7\%                     & 78.2 & 1272   & 90.4\%                     & 26.7 & 5301   & 83.5\%                     & -0.5        & -64.5\%        \\
\bottomrule 
\end{tabular}
}
\caption{Performance of \textit{AdaptThink}-1.5B using different $\delta$ value. 
}
\label{tab:ablation}
\end{table*}

\begin{itemize}[itemsep=0pt, leftmargin=*]

\item\textbf{$\text{DPO}_\textit{Shortest}$} constructs preference data by sampling multiple responses for each problem in the training dataset and pairing the shortest correct response and the longest responses, then uses DPO~\cite{dpo} to finetune the model.

\item\textbf{OverThink}~\cite{chen2025overthink} first constructs preference data by taking the original long-thinking response for each training problem as the negative example and retaining the first two attempts that arrive at the correct answer in the thinking as the positive example, and then uses SimPO~\cite{simpo} to alleviate models' overthinking behaviors.

\item\textbf{DAST}~\cite{shen2025dast} first constructs preference data by ranking pre-sampled responses with a length-based reward function, and then employs SimPO to finetune the model.

\item\textbf{O1-Pruner}~\cite{luo2025o1pruner}  first estimates the reference model's performance through pre-sampling and then uses off-policy RL-style fine-tuning to encourage the model to generate shorter reasoning processes under accuracy constraints.

\item\textbf{TLMRE}~\cite{arora2025training} incorporates a length-based penalty in on-policy RL to incentivize the model to produce shorter responses.

\item\textbf{ModelMerging}~\cite{wu2025unlocking} reduces the response length of a reasoning model by weightedly averaging its weights with a non-reasoning model (i.e., Qwen-2.5-Math-1.5B/7B). 

\item\textbf{$\text{RFT}_\textit{MixThinking}$} (Reject Fine-tuning) first samples multiple responses for each training problem $x$ using both \textit{Thinking} and \textit{NoThinking}, then selects (1) correct \textit{NoThinking} responses if the instance-level pass rate $\bar{R}_\text{nothink}(x) \!\ge\! \bar{R}_\text{think}(x)$ and (2) correct \textit{Thinking} responses if $\bar{R}_\text{nothink}(x) \!<\! \bar{R}_\text{think}(x)$, and uses these selected responses to finetune the model.

\end{itemize}
For fair comparison, we re-implement all these baselines using DeepScaleR dataset.

\subsection{Main Results}

Table~\ref{tab:main} presents the evaluation results of different methods on GSM8K, MATH500, and AIME 2024. Compared to the original 1.5B and 7B models, \textit{AdaptThink} reduces the average response length by 53.0\% and 40.1\%, respectively, while also improves the average accuracy by 2.4\% and 2.3\%,  demonstrating that \textit{AdaptThink} enables significantly more efficient reasoning without compromising and even enhancing model performance. Moreover, \textit{AdaptThink} outperforms most baselines—all of which optimize response length within the \textit{Thinking} mode—in terms of both accuracy and length reduction. It also achieves the best average results, highlighting the effictiveness of adaptive thinking-mode selection as a novel paradigm for achieving efficient reasoning. 

For the methods that involve both \textit{Thinking} and \textit{NoThinking} modes, we additionally report the ratio of responses generated in \textit{NoThinking} mode (i.e., $\text{Ratio}_\textit{NT}$ in Table~\ref{tab:main}). As shown in the table, \textit{AdaptThink} produces more \textit{NoThinking} responses for easier test sets (i.e., GSM8K and MATH500) while employing \textit{Thinking} mode more frequently for challenging test sets (i.e., AIME 2024). See Appendix~\ref{sec:case_study} for detailed cases.
A similar trend is also observed within the five difficulty levels of MATH500 (Figure~\ref{fig:nothinking_ratio_math500_7B} and~\ref{fig:nothinking_ratio_math500_1.5B}), where \textit{AdaptThink} predominantly selects the \textit{NoThinking} mode for the easiest Level-1 problems, and progressively increases the use of \textit{Thinking} as the difficulty level rises. Meanwhile, compared to the original models using only \textit{Thinking} or \textit{NoThinking} mode, \textit{AdaptThink} consistently achieves higher accuracy across most difficulty levels. These findings suggest that \textit{AdaptThink} has successfully taught the model to adaptively choose an appropriate thinking mode based on problem difficulty, achieving a better balance between efficiency and performance. 

\begin{figure*}
    \centering
    \includegraphics[width=\linewidth]{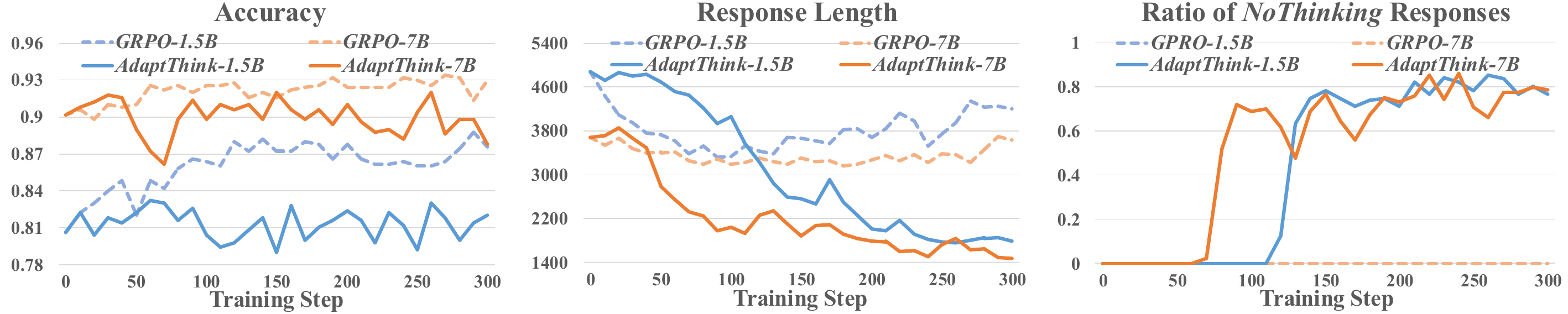}
    \caption{Comparison of accuracy, response length, and the ratio of \textit{NoThinking} response on MATH500 between $\textit{AdaptThink}$ and naive GPRO at different training steps.}
    \label{fig:compare_gpro}
\end{figure*}

\subsection{More Analyses}
\label{sec:analyses}

\paragraph{Effect of $\delta$.}
To show the effect of $\delta$ in our advantage function $A(x, y)$, we implement \textit{AdaptThink} with different $\delta$ values on the 1.5B model and summarize the evaluation results in Table~\ref{tab:ablation}. As $\delta$ increases, the proportion of \textit{NoThinking} responses progressively rises, resulting in a corresponding reduction in the average response length. However, the gain in accuracy also gradually decreases at the same time. This indicates that $\delta$ serves as a control parameter for the trade-off between reasoning efficiency and accuracy improvement. Notably, even when $\delta\!=\!0$, the model chooses \textit{NoThinking} for over 50\% of problems in GSM8K and MATH500, implying that \textit{NoThinking} may possess an inherent advantage over \textit{Thinking} when addressing relatively straightforward problems. Furthermore, for most $\delta$ values, \textit{AdaptThink} consistently achieves notable reduction in response length and also improves accuracy, which underscores its robustness.

\paragraph{Effect of importance sampling.} To demonstrate the effect of importance sampling, we compare \textit{AdaptThink} with naive GRPO that samples directly from $\pi_{\theta_\text{old}}(\cdot|x)$ during training. As shown in Figure~\ref{fig:compare_gpro}, because $\pi_{\theta_\text{old}}(\cdot|x)$ is initially unable to produce \textit{NoThinking} samples, GRPO can only learn from \textit{Thinking} samples and focus on improving accuracy throughout the training process. The response length of GRPO on MATH500 decreases only to around 3,500 (by eliminating overlong responses that would be truncated and receive no reward), after which it gradually increases. In contrast, our importance sampling strategy enables \textit{AdaptThink} to learn from both \textit{Thinking} and \textit{NoThinking} samples at each training step. As the model gradually learn to generate more \textit{NoThinking} responses for simple problems, the response length eventually decreases to below 2,000 tokens.

\begin{table}[!t]
\centering
\resizebox{\linewidth}{!}{
\begin{tabular}{l|cc}
\toprule
Method                       & $\text{Ratio}_\textit{IT}$ & Length \\ \midrule
\multicolumn{3}{l}{\textit{DeepSeek-R1-Distill-Qwen-1.5B}}                \\ \midrule
Final Solutions from $\text{Original}_\textit{Thinking}$  & 9.5\%               & 1799                   \\
$\text{Original}_\textit{NoThinking}$  & 8.2\%               & 665                   \\
\textit{AdaptThink}      & 7.9\%               & 826                   \\ \midrule
\multicolumn{3}{l}{\textit{DeepSeek-R1-Distill-Qwen-7B}}                  \\ \midrule
Final Solutions from $\text{Original}_\textit{Thinking}$  & 0.7\%               & 321                   \\
$\text{Original}_\textit{NoThinking}$  & 0.9\%               & 341                   \\
\textit{AdaptThink}      & 4.2\%               & 426                   \\ \bottomrule
\end{tabular}
}
\caption{For the test cases where \textit{AdaptThink} chooses \textit{NoThinking}, we compare the implicit thinking ratio ($\text{Ratio}_\textit{IT}$) and average length of three scenarios.}
\label{tab:implicit_thinking_ratio}
\end{table}
\paragraph{Implicit thinking ratio.} A potential concern for \textit{AdaptThink} is that RL may activate thinking features (e.g., reflection) within \textit{NoThinking} mode (similar to DeepSeek-R1-Zero) and produce many implicit thinking responses. To allay this concern, we examine the test cases where \textit{AdaptThink} chooses the \textit{NoThinking} mode. For these cases, we collect (1) \textit{NoThinking} responses from \textit{AdaptThink}, (2) \textit{NoThinking} responses from the original reasoning model, and (3) the final solution part of \textit{Thinking} responses from the original model. We compare the ratio of implicit thinking responses (denoted by $\text{Ratio}_\textit{IT}$) across these three scenarios by detecting whether some representative keywords for thinking (e.g, \textit{``Wait''} and \textit{``Alternatively''}) appear in the solutions. We also compare the average length of these solutions. As presented in Table~\ref{tab:implicit_thinking_ratio}, \textit{AdaptThink} only slightly increases the implicit thinking ratio and response length for the 7B model. To entirely eliminate such behavior, one possible approach is to assign zero reward to implicit thinking samples during RL training. 

\begin{table}[!t]
\centering
\resizebox{\linewidth}{!}{
    \setlength{\tabcolsep}{3pt}
\begin{tabular}{l|ccccc}
\toprule
\multirow{2}{*}{Method}               & \multicolumn{5}{c}{MMLU}                                                  \\
                                      & Acc  & Length & $\text{Ratio}_\textit{NT}$ & $\Delta$Acc & $\Delta$Length \\ \midrule
\multicolumn{6}{l}{\textit{DeepSeek-R1-Distill-Qwen-1.5B}}                                                        \\ \midrule
$\text{Original}_\textit{Thinking}$   & 35.7 & 1724   & 0.00\%                      & -           & -              \\
$\text{Original}_\textit{NoThinking}$ & 20.6 & 208    & 100.00\%                    & -15.1       & -87.9\%        \\
\textit{AdaptThink}                & 42.2 & 1055   & 16.43\%                     & +6.5        & -38.8\%        \\ \midrule
\multicolumn{6}{l}{\textit{DeepSeek-R1-Distill-Qwen-7B}}                                                          \\ \midrule
$\text{Original}_\textit{Thinking}$   & 63.4 & 1257   & 0.00\%                      & -           & -              \\ 
$\text{Original}_\textit{NoThinking}$ & 51.2 & 128    & 100.00\%                    & -12.2       & -89.8\%        \\
\textit{AdaptThink}                & 63.6 & 856    & 16.41\%                     & +0.2        & -31.9\%        \\ \bottomrule
\end{tabular}
}
\caption{The performance of \textit{AdaptThink} on the out-of-distribute test set MMLU.}
\label{tab:ood}
\end{table}
\paragraph{Generalizability to OOD scenario.} To assess the ability of \textit{AdaptThink} to generalize in out-of-distribution scenarios, we conduct an evaluation on MMLU, which contains 14K multi-choice questions and covers 57 diverse domains, distinct from our training data in question format and subjects. As shown in Table~\ref{tab:ood}, \textit{AdaptThink} reduces average response length by more than 30\% by producing \textit{NoThinking} responses for about 16\% of the problems (see Figure~\ref{case:easy_mmlu} for cases), while achieving higher accuracy than the original models.

\section{Conclusion}
In this work, we first demonstrate the advantages of \textit{NoThinking} over \textit{Thinking} in both performance and efficiency for relatively simple tasks. Motivated by this, we propose \textit{AdaptThink}, a novel RL algorithm to enable reasoning models to adaptively select the optimal thinking mode based on problem difficulty. Experiments show that \textit{AdaptThink} significantly reduces inference costs and further improves model performance, highlighting the promise of adaptive thinking-mode selection for advancing the trade-off between reasoning quality and efficiency.

\section{Limitation}
We discuss several limitations of our work in this section: 
(1) Due to limited computational resources, we only conduct our experiments on 1.5B and 7B models. Nevertheless, these experiments still demonstrate the efficacy of \textit{AdaptThink} across different model sizes. (2) Similar to most previous open-sourced works, we only train our models using mathematical datasets because they are easy to obtain and can offer accurate, verifiable rewards. Though our evaluation on MMLU shows \textit{AdaptThink} models can well generalize to OOD scenarios, we believe they can achieve better results if more training datasets with verifiable rewards for general domains are available. 

\section{Ethical Considerations}
All the models and datasets used in this work are publicly published with permissible licenses. 
\bibliography{custom}

\newpage
\appendix
\onecolumn
\begin{figure*}[!t]
    \centering
    \includegraphics[width=\linewidth]{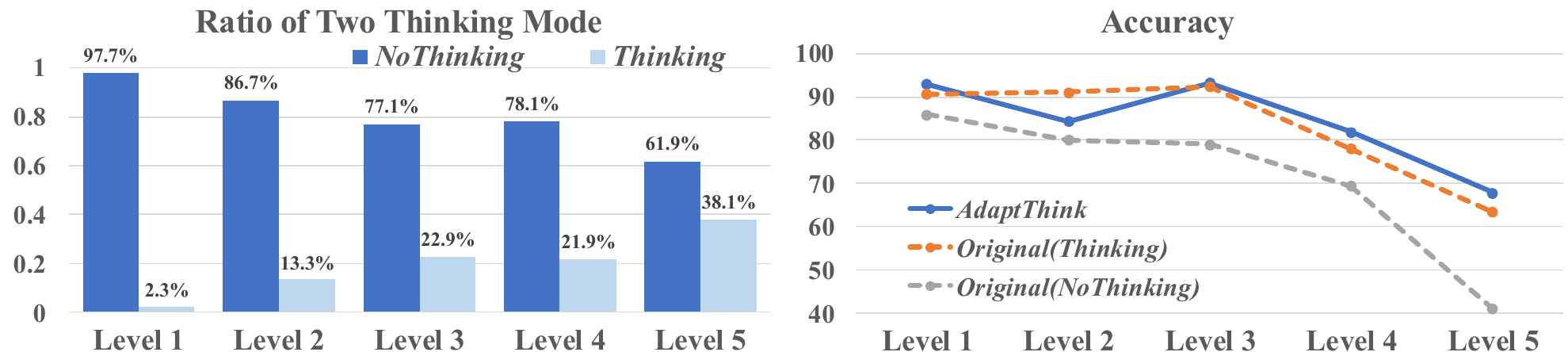}
    \caption{Left: The ratio that \textit{AdaptiveThink}-1.5B choose \textit{Thinking} or \textit{NoThinking} across different MATH levels. Right: Comparison of accuracy between \textit{AdaptiveThink}-1.5B and the DeepSeek-R1-Distill-Qwen-1.5B using \textit{Thinking} and \textit{NoThinking} across different MATH levels.}
    \label{fig:nothinking_ratio_math500_1.5B}
\end{figure*}

\section{Case Study}
\label{sec:case_study}
We show several cases of \textit{AdaptThink} in Figure~\ref{case:easy_gsm8k},~\ref{case:esay_math500},~\ref{case:easy_mmlu}, and ~\ref{case:hard_aime} to illustrate its ability of adaptive thinking-mode selection. As demonstrated in Figure~\ref{case:easy_gsm8k},~\ref{case:esay_math500}, and ~\ref{case:easy_mmlu}, when faced with relatively simple problems, DeepSeek-R1-Distill-Qwen-7B costs a large number of tokens in thinking, which often contains many unnecessary steps and redundant attempts. In contrast, our \textit{AdaptThink}-7B model adaptively chooses \textit{NoThinking} mode and directly produces a concise final solution, even if the input problem differs from our training data in both the question format and the subject. On the hard hand, the case in Figure~\ref{case:hard_aime} indicates that, when the input problem is difficult, \textit{AdaptThink}-7B is also able to automatically employ \textit{Thinking} to ensure its accuracy.

\begin{figure*}
\centering
\begin{tcolorbox}[size=title,opacityfill=0.1]
\noindent
\faUser\ \textbf{Question (GSM8K)}: Two cars are driving on a highway.  The first car is traveling at an average speed of 60 miles per hour when the second car passes it at an average speed of 70 miles per hour.  If both cars continue on the highway at the same speed, how many miles will separate them after 2 hours? \\

\faCogs\ \textbf{DeepSeek-R1-Distill-Qwen-7B}:

\texttt{<think>}Okay, so I have this problem here about two cars on a highway. Let me try to understand it step by step. The first car is going at an average speed of $60$ miles per hour. Then, the second car passes the first car, and it's going at $70$ miles per hour. The question is asking, if both cars keep going at the same speed, how many miles apart will they be after $2$ hours. Hmm, okay. So, let me visualize this. There's a highway, and both cars are moving in the same direction. The first car is ahead, going $60\,\text{mph}$, and the second car catches up to it, going $70\,\text{mph}$. So, the second car is faster, right? It's going $10\,\text{mph}$ faster than the first car. Wait, so the second car is approaching the first car at a rate of 10 miles per hour. That makes sense because 70 minus 60 is 10. So, the gap between them is closing at 10 mph. But the question is asking, after 2 hours, how far apart will they be. Hmm, wait a second. If the second car is catching up, the distance between them is decreasing. But the question says, how many miles will separate them after 2 hours. So, is it possible that the second car overtakes the first car before 2 hours, and then the distance becomes positive again? Or maybe they're both moving in the same direction, so the second car is just gaining on the first car, but after 2 hours, the distance between them is how much? Wait,
\textbf{[...](omit 2679 tokens)} 
so, I think I'm confident that the answer is $20$ miles.\texttt{</think>} \textbf{Final Answer:} 
The number of miles that will separate them after $2$ hours is $\boxed{20}$. The first car is traveling at an average speed of $60$ miles per hour. The second car passes it at an average speed of $70$ miles per hour. Both cars continue on the highway at the same speed. We need to determine how many miles will separate them after $2$ hours. 1. The relative speed between the second car and the first car is the difference in their speeds: $70\,\text{mph} - 60\,\text{mph} = 10\,\text{mph}$. 2. The separation between the two cars after time $t$ hours is given by the relative speed multiplied by time: $\text{Separation} = 10\,\text{mph} \times t\,\text{hours}$. 3. After $2$ hours, the separation will be: $\text{Separation} = 10\,\text{mph} \times 2\,\text{hours} = 20\,\text{miles}$. Thus, the number of miles that will separate them after $2$ hours is $\boxed{20}$. \textbf{(3303 tokens)}
\
\\

{\color[HTML]{2EA121}\faCogs\ \textbf{AdaptThink-7B} (ours)}: \\
\texttt{<think></think>}To determine how many miles will separate the two cars after 2 hours, let's break down the problem step by step. 1. \textbf{Understand the Speeds:} The first car is traveling at $60$ miles per hour (mph). The second car is traveling at $70$ mph. 2. \textbf{Determine the Relative Speed:} Since both cars are moving in the same direction, the relative speed between them is the difference in their speeds. Relative speed $=~$ Speed of second car $-$ Speed of first car $= 70~\mathrm{mph} - 60~\mathrm{mph} = 10~\mathrm{mph}$. 3. \textbf{Calculate the Distance Separated After 2 Hours:} Distance $=$ Speed $\times$ Time. Here, the relative speed is $10~\mathrm{mph}$, and the time is $2$ hours. So, Distance $= 10~\mathrm{mph} \times 2~\mathrm{hours} = 20~\mathrm{miles}$. Therefore, after 2 hours, the two cars will be $\boxed{20}$ miles apart. \textbf{(222 tokens)}

\end{tcolorbox}
\caption{For a simple math problem from GSM8K, DeepSeek-R1-Distill-Qwen-7B costs about 3000 tokens in thinking, which contains many unnecessary steps and redundant attempts. In contrast, our \textit{AdaptThink}-7B adaptively chooses \textit{NoThinking} mode and directly produces a concise final solution.}
\label{case:easy_gsm8k}
\end{figure*}
\begin{figure*}
\centering
\begin{tcolorbox}[size=title,opacityfill=0.1]
\noindent
\faUser\ \textbf{Question (MATH500)}: In how many ways can 8 people sit around a round table if 3 of the people -- Pierre, Rosa, and Thomas -- all want to sit together?  (Two seatings are considered the same if one is a rotation of the other.) \\

\faCogs\ \textbf{DeepSeek-R1-Distill-Qwen-7B}:

\texttt{<think>}Alright, so I have this problem where I need to figure out how many ways 8 people can sit around a round table, but there's a condition: Pierre, Rosa, and Thomas all want to sit together. Hmm, okay. Let me think about this step by step.
First, I remember that when arranging people around a round table, the number of arrangements is different from arranging them in a straight line because rotations of the same arrangement are considered identical. For a round table with $n$ people, the number of distinct seating arrangements is $(n-1)!$ instead of $n!$. That's because fixing one person's position accounts for the rotations, and then arranging the others relative to that fixed position.
But in this case, there's a condition: Pierre, Rosa, and Thomas must all sit together. So, how does that affect the calculation?
I think the way to handle this is to treat the group of Pierre, Rosa, and Thomas as a single entity or "block." If we consider them as one block, then we're effectively arranging this block along with the other 5 people. So, that makes a total of 6 "blocks" to arrange around the table.
Wait, hold on. If we have 8 people and we're grouping 3 of them together, how many people are left? Let me subtract: $8 - 3 = 5$. So, yes, we have 5 individuals plus the block, totaling 6 units to arrange around the table.
Now, since it's a round table, the number of ways to arrange these 6 units is $(6-1)! = 5! = 120$. That's because, again, fixing one position accounts for the rotational symmetry.
But wait, \textbf{[...](omit 9138 tokens)}
Therefore, I think the answer is 720.
\texttt{</think>}
\textbf{Final Answer:}
The number of ways is $\boxed{720}$.
To determine the number of ways 8 people can sit around a round table with the condition that Pierre, Rosa, and Thomas must sit together, we can treat them as a single block. This reduces the problem to arranging 6 entities (the block and the other 5 people) around a circular table.
Treat the group as a single block: This gives us 6 entities (the block and 5 individuals). Arrange these 6 entities around a circular table: The number of distinct arrangements of 6 entities around a circular table is $(6 - 1)! = 5!$. Arrange the people within the block: The 3 people in the block can be arranged among themselves in $3!$ ways.
Thus, the total number of arrangements is the product of these two results:
\(
(6 - 1)! \times 3! = 5! \times 3! = 120 \times 6 = 720
\)
Therefore, the number of ways 8 people can sit around a round table with Pierre, Rosa, and Thomas sitting together is $\boxed{720}$. \textbf{(9812 tokens)}
\
\\

{\color[HTML]{2EA121}\faCogs\ \textbf{AdaptThink-7B} (ours)}: \\
\texttt{<think></think>}To determine the number of ways 8 people can sit around a round table with the condition that Pierre, Rosa, and Thomas must all sit together, we can treat the group of three as a single entity or "block." Here's the step-by-step solution: 
\textbf{1. Treat Pierre, Rosa, and Thomas as a single block:} This reduces the problem to arranging 6 entities around a round table: the block and the other 5 individuals. 
\textbf{2. Arrange the 6 entities around the table:} For circular arrangements, the number of ways to arrange \( n \) entities is \( (n-1)! \). So, for 6 entities, it's \( (6-1)! = 5! = 120 \) ways. 
\textbf{3. Arrange Pierre, Rosa, and Thomas within their block:} The three people can be arranged among themselves in \( 3! = 6 \) ways. 
\textbf{4. Multiply the two results together:} The total number of arrangements is the product of the arrangements of the blocks and the internal arrangements of the group: \( 5! \times 3! = 120 \times 6 = 720 \).
So, the number of ways 8 people can sit around a round table with Pierre, Rosa, and Thomas all sitting together is \( \boxed{720} \). \textbf{(297 tokens)}

\end{tcolorbox}
\caption{For a relatively simple math problem from MATH500, DeepSeek-R1-Distill-Qwen-7B costs about 9000 tokens in thinking. In contrast, our \textit{AdaptThink}-7B adaptively chooses \textit{NoThinking} mode and directly produces a concise final solution, using only 297 tokens.}
\label{case:esay_math500}
\end{figure*}
\begin{figure*}
\centering
\begin{tcolorbox}[size=title,opacityfill=0.1]
\noindent
\faUser\ \textbf{Question (MMLU)}: A proposed mechanism for a reaction is as follows:  
\(\mathrm{NO_2 + F_2 \rightarrow NO_2F + F}\) (Slow step).  
\(\mathrm{F + NO_2 \rightarrow NO_2F}\) (Fast step).  
What is the order of the overall reaction?  
A) Zero order.  
B) First order.  
C) Second order.  
D) Third order. 
Output the correct option (one of ABCD) in "$\backslash$boxed\{\}", such as "$\backslash$boxed\{A\}".\\

\faCogs\ \textbf{DeepSeek-R1-Distill-Qwen-7B}:

\texttt{<think>}Okay, so I have this chemistry problem here about reaction mechanisms and determining the overall order of the reaction. Hmm, let me try to figure this out step by step.
The problem says the proposed mechanism is:  
1. $\mathrm{NO}_2 + \mathrm{F}_2 \rightarrow \mathrm{NO}_2\mathrm{F} + \mathrm{F}$ (slow step)  
2. $\mathrm{F} + \mathrm{NO}_2 \rightarrow \mathrm{NO}_2\mathrm{F}$ (fast step)
And the question is asking for the order of the overall reaction. The options are A. Zero order, B. First order, C. Second order, D. Third order.
Alright, I remember that the overall order of a reaction is determined by the sum of the orders of each reactant in the rate law. But wait, the rate law isn't always given. In this case, the mechanism is provided, so I need to use that to figure out the rate law.
First, in a reaction mechanism, the slow step is the rate-determining step, right? That means the rate law is based on that step. So I should focus on the first step: $\mathrm{NO}_2 + \mathrm{F}_2 \rightarrow \mathrm{NO}_2\mathrm{F} + \mathrm{F}$.
Looking at the slow step, the reactants are $\mathrm{NO}_2$ and $\mathrm{F}_2$. So the rate law would be $\mathrm{rate} = k[\mathrm{NO}_2][\mathrm{F}_2]^m$, where $m$ is the order with respect to F$_2$. But wait, in the slow step, F$_2$ is a reactant, but in the second step, there's an F that's produced. Hmm, I think that in the overall reaction, the intermediates (like F in this case) should be canceled out to get the rate law in terms of the initial reactants. Hold on, 
\textbf{[...](omit 3887 tokens)}
Therefore, the answer is C. Second order.
\textbf{Final Answer}  
$\boxed{C}$
\texttt{</think>}
The proposed mechanism for the reaction is as follows:
1. $\mathrm{NO}_2 + \mathrm{F}_2 \rightarrow \mathrm{NO}_2\mathrm{F} + \mathrm{F}$ (slow step)  
2. $\mathrm{F} + \mathrm{NO}_2 \rightarrow \mathrm{NO}_2\mathrm{F}$ (fast step)
The overall reaction is determined by combining the two steps:
$\mathrm{NO}_2 + \mathrm{F}_2 + \mathrm{F} + \mathrm{NO}_2 \rightarrow \mathrm{NO}_2\mathrm{F} + \mathrm{F} + \mathrm{NO}_2\mathrm{F}$.
Simplifying, we get:
$2\,\mathrm{NO}_2 + \mathrm{F}_2 \rightarrow 2\,\mathrm{NO}_2\mathrm{F}$.
The rate law is determined by the slow step, which is the rate-determining step. The rate law for the slow step is:
$\mathrm{rate} = k[\mathrm{NO}_2][\mathrm{F}_2]$.
Since $\mathrm{F}$ is an intermediate, we use the steady-state approximation to express $[\mathrm{F}]$ in terms of $[\mathrm{NO}_2]$ and $[\mathrm{F}_2]$. From the fast step, we find:
$k_1[\mathrm{NO}_2][\mathrm{F}_2] = k_2[\mathrm{F}][\mathrm{NO}_2]$.
Solving for $[\mathrm{F}]$:
$[\mathrm{F}] = \frac{k_1}{k_2}[\mathrm{F}_2]$.
Substituting this back into the rate law confirms that the rate is still first order in $[\mathrm{NO}_2]$ and first order in $[\mathrm{F}_2]$. Therefore, the overall order of the reaction is:
$1 + 1 = 2$.
Thus, the order of the overall reaction is second order.
$\boxed{C}$ \textbf{(4709 tokens)}
\
\\

{\color[HTML]{2EA121}\faCogs\ \textbf{AdaptThink-7B} (ours)}: \\
\texttt{<think></think>}To determine the order of the overall reaction, we analyze the given mechanism:
\textbf{1. Slow Step (Rate-Determining Step):} 
\(
\text{NO}_2 + \text{F}_2 \rightarrow \text{NO}_2\text{F} + \text{F}
\)
The rate law for this step is: \(
\text{Rate} = k[\text{NO}_2][\text{F}_2].
\).
\textbf{2. Fast Step:} 
\(
\text{F} + \text{NO}_2 \rightarrow \text{NO}_2\text{F}.
\)
Since this step is fast, it does not affect the overall rate law. The overall rate law is determined by the slow step.
Combining the rate laws from the slow step, the overall reaction order is the sum of the exponents in the rate law: the order with respect to \(\text{NO}_2\) is 1, and the order with respect to \(\text{F}_2\) is 1.
Thus, the overall reaction order is \(1 + 1 = 2\).
The correct answer is \(\boxed{C}\). \textbf{(263 tokens)}

\end{tcolorbox}
\caption{Even for out-of-distribution problems from MMLU, \textit{AdaptThink}-7B is still able to generate \textit{NoThinking} response if the problem is relatively simple.}
\label{case:easy_mmlu}
\end{figure*}
\begin{figure*}
\centering
\begin{tcolorbox}[size=title,opacityfill=0.1]
\noindent
\faUser\ \textbf{Question (AIME 2024)}: Find the largest possible real part of \((75+117i)z+\frac{96+144i}{z}\) where \(z\) is a complex number with \(|z|=4\).\\

\faCogs\ \textbf{DeepSeek-R1-Distill-Qwen-7B}:

\texttt{<think>}Alright, so I have this problem here: I need to find the largest possible real part of the expression $(75 + 117i)z + \frac{96 + 144i}{z}$ where $z$ is a complex number with $|z| = 4$. Hmm, okay, let me try to figure this out step by step. 
First, I remember that for complex numbers, the real part of an expression is the sum of the real parts of each term. So, I can probably break this down into two parts: the real part of $(75 + 117i)z$ and the real part of $\frac{96 + 144i}{z}$. Then, 
\textbf{[...](omit 3099 tokens)}
I think that solidifies the answer. I don't see any mistakes in the reasoning.
\textbf{Final Answer} $\boxed{540}$
\texttt{</think>}
Given the problem of finding the largest possible real part of the expression $(75 + 117i)z + \frac{96 + 144i}{z}$ where $z$ is a complex number with $|z| = 4$, we start by expressing $z$ as $z = 4e^{i\theta}$. This allows us to rewrite the expression in terms of $\theta$.
First, we substitute $z = 4e^{i\theta}$ and $\frac{1}{z} = \frac{1}{4}e^{-i\theta}$ into the expression: $(75 + 117i) \cdot 4e^{i\theta} + (96 + 144i) \cdot \frac{1}{4}e^{-i\theta}$.
This simplifies to $(300 + 468i)e^{i\theta} + (24 + 36i)e^{-i\theta}$.
Expanding each term using Euler's formula $e^{i\theta} = \cos\theta + i\sin\theta$ and $e^{-i\theta} = \cos\theta - i\sin\theta$, we separate the real and imaginary parts. After combining the real parts, we get $324\cos\theta - 432\sin\theta$.
To find the maximum value of this expression, we recognize it as a linear combination of sine and cosine, which can be written in the form $A\cos\theta + B\sin\theta$. The maximum value of such an expression is given by the amplitude $R = \sqrt{A^2 + B^2}$.
Calculating $R$: $R = \sqrt{324^2 + (-432)^2} = \sqrt{104976 + 186624} = \sqrt{291600} = 540$.
Thus, the largest possible real part of the given expression is $\boxed{540}$. \textbf{(3771 tokens)}
\
\\

{\color[HTML]{2EA121}\faCogs\ \textbf{AdaptThink-7B} (ours)}: \\
\texttt{<think>}Okay, so I have this problem here: I need to find the largest possible real part of the expression \( (75 + 117i)z + \frac{96 + 144i}{z} \), where \( z \) is a complex number with \( |z| = 4 \). Hmm, okay. Let me try to figure this out step by step.
First, I know that \( z \) is a complex number with magnitude 4. That means 
\textbf{[...](omit 3003 tokens)}
Therefore, the largest possible real part is \( \boxed{540} \).
\texttt{</think>}
Given the expression \( (75 + 117i)z + \frac{96 + 144i}{z} \) where \( z \) is a complex number with \( |z| = 4 \), we need to find the largest possible real part.
First, we express \( z \) as \( 4e^{i\theta} \) where \( \theta \) is the argument of \( z \). Then, we rewrite the expression in terms of \( \theta \): \( E = (75 + 117i)4e^{i\theta} + \frac{96 + 144i}{4}e^{-i\theta} \).
Simplifying the constants, we get: \( (75 + 117i)4 = 300 + 468i \) and \( \frac{96 + 144i}{4} = 24 + 36i \).
Thus, the expression becomes: \( E = (300 + 468i)e^{i\theta} + (24 + 36i)e^{-i\theta} \).
Using Euler's formula, we expand and separate into real and imaginary parts: \( (300 + 468i)(\cos\theta + i\sin\theta) = (300\cos\theta - 468\sin\theta) + i(300\sin\theta + 468\cos\theta) \) and \( (24 + 36i)(\cos\theta - i\sin\theta) = (24\cos\theta + 36\sin\theta) + i(-24\sin\theta + 36\cos\theta) \).
Adding the real parts together: \( 300\cos\theta - 468\sin\theta + 24\cos\theta + 36\sin\theta = 324\cos\theta - 432\sin\theta \).
The expression \( 324\cos\theta - 432\sin\theta \) can be written as \( R\cos(\theta + \phi) \) where \( R = \sqrt{324^2 + (-432)^2} \). Calculating \( R \): \( 324^2 = 104976 \), \( 432^2 = 186624 \), \( R = \sqrt{104976 + 186624} = \sqrt{291600} = 540 \).
Thus, the largest possible real part is \( \boxed{540} \). \textbf{(3743 tokens)}

\end{tcolorbox}
\caption{For a chanlleging problems from AIME 2024, \textit{AdaptThink}-7B is able to employs \textit{Thinking} to solve it, instead of directly generates the final solution.}
\label{case:hard_aime}
\end{figure*}

\end{document}